\documentclass[runningheads]{llncs}
\usepackage{graphicx}
\usepackage{comment}
\usepackage{amsmath,amssymb} 
\usepackage{color}

\usepackage{epsfig}

\usepackage{algorithm}
\usepackage{algpseudocode}

\usepackage{multirow}
\usepackage[misc]{ifsym}
\usepackage{hyperref}

\begin{document}
	\pagestyle{headings}
	\mainmatter
	\def\ECCVSubNumber{2183}  
	\title{AE TextSpotter: Learning Visual and Linguistic Representation for Ambiguous Text Spotting}
	
	
	\titlerunning{AE TextSpotter}
	%
	\author{Wenhai Wang\inst{1} \and Xuebo Liu\inst{2} \and Xiaozhong Ji\inst{1} \and Enze Xie\inst{3} \and Ding Liang\inst{2} \\ Zhibo Yang\inst{4} \and Tong Lu\inst{1,}$^{\textrm{\Letter}}$ \and Chunhua Shen\inst{5} \and Ping Luo\inst{3}}
	\authorrunning{W. Wang et al.}
	
	\institute{$^{1}$National Key Lab for Novel Software Technology, Nanjing University \\
	$^{2}$SenseTime Research \quad
	$^{3}$The University of Hong Kong \\
	$^{4}$Alibaba-Group \quad
	$^{5}$The University of Adelaide\\
	\tt\small wangwenhai362@smail.nju.edu.cn, \{liuxuebo, liangding\}@sensetime.com, shawn\_ji@163.com, xieenze@hku.hk, zhibo.yzb@alibaba-inc.com, lutong@nju.edu.cn, chunhua.shen@adelaide.edu.au, pluo@cs.hku.hk}
	
	%
	\maketitle

\begin{abstract}
	Scene text spotting aims to detect and recognize the entire word or sentence with multiple characters in natural images.
	It is still challenging because ambiguity often occurs 
	when the spacing between characters is large or the characters are evenly spread in multiple rows and columns, making many visually plausible groupings of the characters (\emph{e.g.} ``BERLIN'' is incorrectly detected as ``BERL'' and ``IN'' in Fig.~\ref{fig:cmp}(c)).
	Unlike previous works that merely employed visual features for text detection, this work proposes a novel text spotter, named Ambiguity Eliminating Text Spotter (AE TextSpotter), which learns both visual and linguistic features to significantly reduce ambiguity in text detection.
	The proposed AE TextSpotter has three important benefits.
	1) The linguistic representation is learned together with the visual representation in a framework. To our knowledge, it is the first time to improve text detection by using a language model.
	2) A carefully designed language module is utilized to reduce the detection confidence of incorrect text lines, making them easily pruned in the detection stage.
	3) Extensive experiments show that AE TextSpotter outperforms other state-of-the-art methods by a large margin. For example, we carefully select a set of extremely ambiguous samples from the IC19-ReCTS dataset, where our approach surpasses other methods by more than 4\%.
\end{abstract}

\section{Introduction}

Text analysis in unconstrained scene images like text detection and text recognition is important in many applications, such as document recognition, license plate recognition, and visual question answering based on texts.
Although previous works~\cite{tian2016ctpn,zhou2017east,liu2018fots,long2018textsnake,li2018psenet,liao2019maskpami} have acquired great success, there are still many challenges to be solved.

\begin{figure*}[t]
	\centering
	\setlength{\fboxrule}{0pt}
	\fbox{\includegraphics[width=0.85\textwidth]{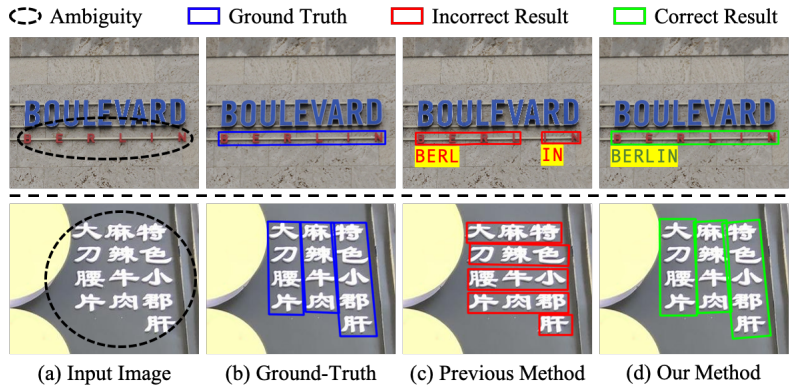}}
	\caption{The detection results of different methods in cases of ambiguity. Top row: an example of large character spacing. Bottom row: an example of juxtaposed text lines. (a) are original images. (b) are ground-truths. (c) are detection results of Mask TextSpotter~\cite{liao2019maskpami}. (d) are our detection results.}
	\label{fig:cmp}
\end{figure*}

This work addresses one of the important challenges, which is reducing the ambiguous bounding box proposals in scene text detection. These ambiguous proposals widely occur when the spacing of the characters of a word is large or multiple text lines are juxtaposed
in different rows or columns 
in an image. For example, as shown in Fig.~\ref{fig:cmp}(c)(d), the characters in an image can be grouped into multiple visually plausible words, making the detection results ambiguous and significantly hindering the accuracy of text detectors and text spotters.

\begin{figure*}[t]
	\centering
	\setlength{\fboxrule}{0pt}
	\fbox{\includegraphics[width=0.95\textwidth]{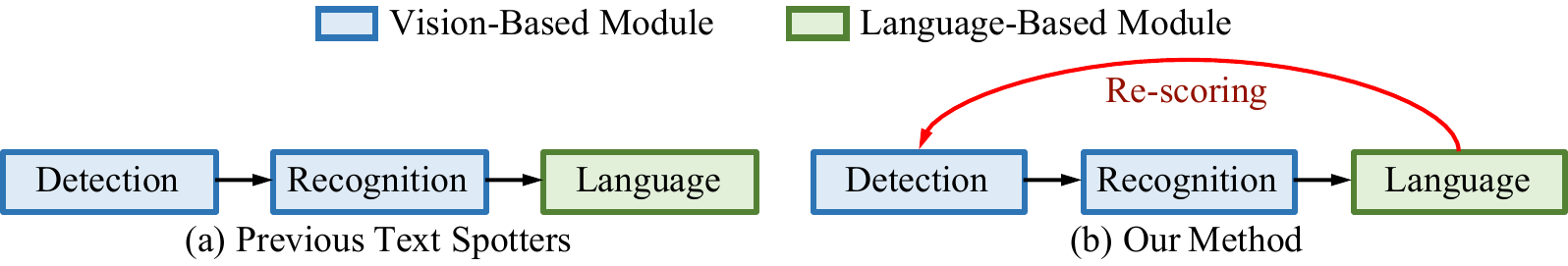}}
	\caption{\textbf{Comparisons} between (a) previous text spotters and (b) our AE TextSpotter. Different from previous methods \cite{liu2018fots,liao2019maskpami,qin2019towards,feng2019textdragon} where 
		the modules of text detection and linguistic knowledge are isolated, our method lifts the language module ahead in text detection to re-score the detection results by learning linguistic representation.
	}
	\label{fig:pipeline}
\end{figure*}

Existing text spotters \cite{liu2018fots,liao2019maskpami,qin2019towards,feng2019textdragon} typically follow a pipeline as shown in Fig.~\ref{fig:pipeline}(a), where a text recognition module is stacked after a text detection module, and linguistic knowledge (\emph{e.g.} lexicon) is employed for text recognition.
However, their language-based modules are isolated from the detection modules, which only detect text lines by learning visual features, leading to ambiguous proposals and hindering the accuracy of the entire systems.
As shown in Fig.~\ref{fig:cmp}(c), these vision-based text detectors are insufficient to detect text lines
correctly in ambiguous samples (\emph{i.e.} horizontal bounding box versus vertical bounding box in this example).

In contrast, unlike existing text detectors that only utilize vision knowledge,
this work incorporates linguistic knowledge into text detection by learning linguistic representation to reduce ambiguous proposals. As shown in Fig.~\ref{fig:pipeline}(b), the re-scoring step is not only the main difference between our approach and the previous methods, but also the key step to remove ambiguity in text detection.
Intuitively, without a natural language model, it is difficult to identify the correct text line in ambiguous scenarios.
For example, prior art \cite{liao2019maskpami} detected incorrect text lines as shown in Fig.\ref{fig:cmp}(c), neglecting the semantic meaning of text lines.
In our approach, the knowledge of natural language can be used to reduce errors as shown in Fig.~\ref{fig:cmp}(d), because the text content may follow the distribution of natural language if the detected text line is correct, otherwise the text content is out of the distribution of natural language. Therefore, the linguistic representation can help refine the detection results.

This paper proposes a novel framework called  Ambiguity Eliminating Text Spotter~(AE TextSpotter) as shown in Fig.~\ref{fig:pipeline}(b), to solve the ambiguity problem in text detection. 
The rationale of our method is to re-score the detection results by using linguistic representation, 
and eliminate ambiguity by filtering out low-scored text line proposals.
Specifically, AE TextSpotter consists of three important components, including detection, recognition and rescoring.

In the detection step, we build a text detection module (TDM) to provide candidates of text lines with high recall, in order to avoid missed detection.
In this step, the number of candidates is very large because of the high-recall rate. 
%
In the step of recognition, we carefully design a character-based recognition module~(CRM), to efficiently recognize the candidate text lines. 
%
In the re-scoring step, we propose a language module (LM) that can learn linguistic representation to re-score the candidate text lines and to eliminate ambiguity, making the text lines that correspond to natural language have higher scores than those not.
%
The three steps above complement each other, enabling the AE TextSpotter to successfully detect and recognize text lines even
in extremely challenging scenarios. 
As shown in Fig.~\ref{fig:cmp}(d), AE TextSpotter can correctly detect text lines when multiple vertical and horizontal text lines are presented.

Our main \textbf{contributions} are summarized as follows. \textbf{1)} The AE TextSpotter identifies the importance of linguistic representation for text detection, in order to solve the text detection ambiguity problem. 
To our knowledge, this is the first time to learn both visual features and linguistic features for text detection in a framework.
\textbf{2)} We design a novel text recognition module, called the character-based recognition module (CRM), to achieve fast recognition of numerous candidate text lines. 
\textbf{3)} 
%
Extensive experiments on several benchmarks demonstrate the effectiveness of AE TextSpotter. For example, AE TextSpotter achieves state-of-the-art performance on both text detection and recognition on IC19-ReCTS \cite{zhang2019icdar}. 
Moreover, we construct a hard validation set for benchmarking \textbf{T}ext \textbf{D}etection \textbf{A}mbiguity, termed TDA-ReCTS, where AE TextSpotter achieves the F-measure of 81.39\% and the 1-NED of 51.32\%, outperforming its counterparts by 4.01\% and 4.65\%.

\section{Related Work}

\textbf{Scene text detection} has been a research hotspot in computer vision for a long period. 
Methods based on deep learning have become the mainstream of scene text detection. Tian \emph{et al.}~\cite{tian2016ctpn} and Liao \emph{et al.}~\cite{liao2017textboxes} successfully adopted the framework of object detection into text detection and achieved good performance on horizontal text detection. After that, many works~\cite{zhou2017east,shi2017seglink,deng2018pixellink,liao2018textboxes++,liu2019pmtd} took the orientation of text lines into consideration and make it possible to detect arbitrary-oriented text lines. Recently, curved text detection attracted increasing attention, and segmentation-based methods~\cite{long2018textsnake,li2018psenet,wang2019efficient,xie2019spcnet} achieved excellent performances over the curved text benchmarks. These methods improve the performance of text detection to a high level, but none of them can deal with the ambiguity problem in text detection. 
In this work,
we introduce linguistic features in the text detection module to solve the text detection ambiguity problem.

\textbf{Scene text recognition} targets to decode a character sequence from variable-length text images. 
CRNN~\cite{shi2016crnn} introduced CTC~\cite{graves2006ctc} in a text recognition model, and made the model trained end-to-end. Following this pipeline, other CTC-based methods~\cite{liu2016star,wang2017gated} were proposed and achieved significant improvement in text recognition. Another direction of text recognition is attention-based encoder and decoder framework~\cite{shi2018aster,cheng2018aon,cheng2017focusing}, which focuses on one position and predict the corresponding character at each time step. These attention-based methods have achieved high accuracy on various benchmarks. However, most existing methods work slowly when recognizing a large number of candidate text lines in an image. In this paper, we design a character-based recognition module to solve this problem.

\textbf{Scene text spotting} aims to detect and recognize text lines simultaneously. 
Li \emph{et al.} \cite{li2017iccv} proposed a framework for horizontal text spotting, which contains a text proposal network
for text detection and an attention-based method for text recognition. 
FOTS~\cite{liu2018fots} presented a method for oriented text spotting by new RoIRotate operation, and significantly improved the performance by jointly training detection and recognition tasks.
Mask TextSpotter~\cite{liao2019maskpami} added character-level supervision to Mask R-CNN~\cite{he2017mask} to simultaneously identity text lines and characters, which is suitable for curved text spotting. 
Qin \emph{et al.}~\cite{qin2019towards} also developed a curved text spotter based on Mask R-CNN, and used a RoI masking operation to extract the feature for curved text recognition.
However, in these methods, linguistic knowledge~(\emph{e.g.} lexicon) is only used in post-processing, so it is isolated from network training.
Different from them, the linguistic representation in our method is learned together with the visual representation in a framework, making it possible to eliminate ambiguity in text detection.

\section{Proposed Method}
\subsection{Overall Architecture}
\label{sec:arch}
\begin{figure*}[t]
	\centering
	\setlength{\fboxrule}{0pt}
	\fbox{\includegraphics[width=0.98\textwidth]{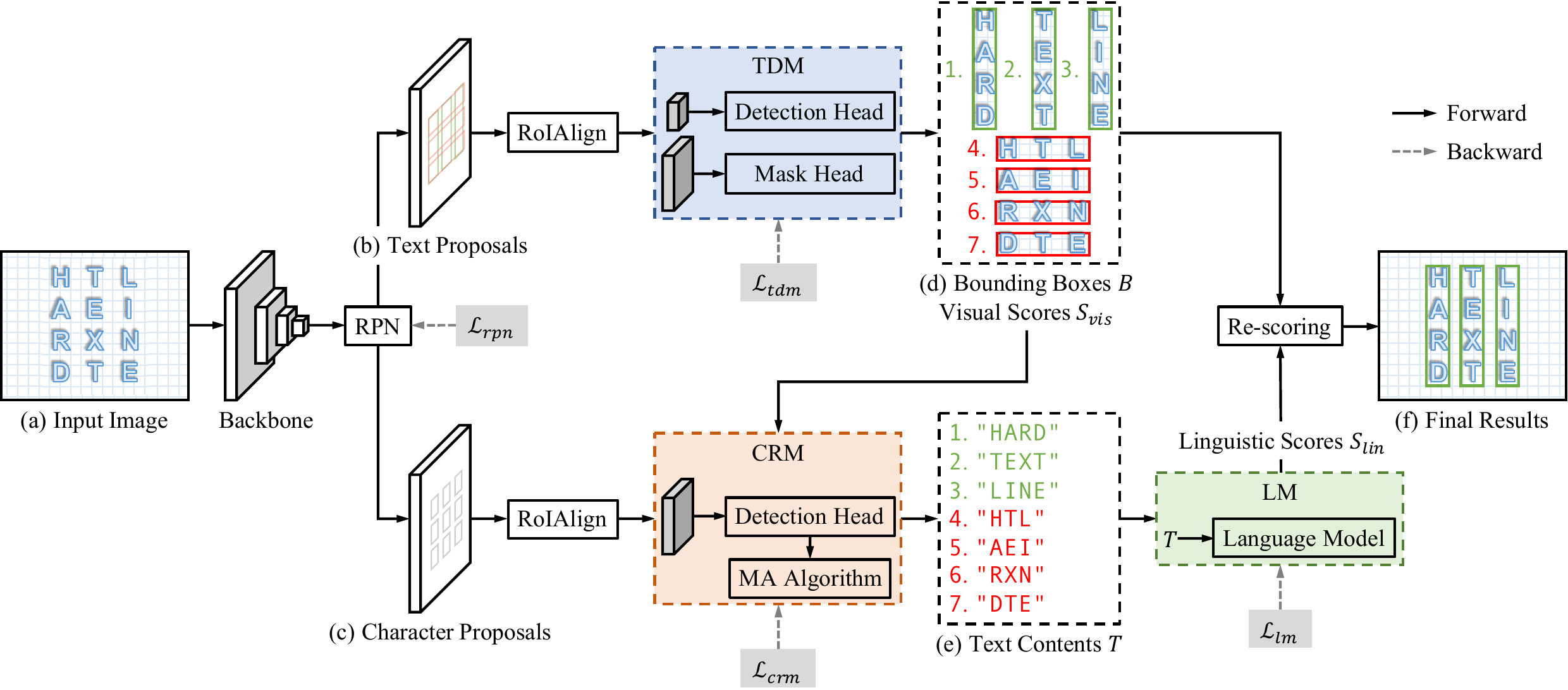}}
	\caption{The overall architecture of AE TextSpotter, which has three key modules: 1) Text Detection Module~(TDM); 2) Character-based Recognition Module~(CRM); 3) Language Module~(LM).
	}
	\label{fig:arch}
\end{figure*}
Fig.~\ref{fig:arch} shows the overall architecture of AE TextSpotter, which consists of two vision-based modules and one language-based module, namely, the text detection module (TDM), the character-based recognition module~(CRM), and the language module~(LM).
Among these modules, TDM and CRM aim to detect the bounding boxes and recognize the content of candidate text lines; 
and LM is applied to lower the scores of incorrect text lines by utilizing linguistic features,
which is the key module to remove ambiguous samples.

In the forward phase, we firstly feed the input image to the backbone network~(\emph{i.e.}, ResNet-50~\cite{he2016deep} + FPN~\cite{lin2017feature})
and Region Proposal Network~(RPN)~\cite{ren2015faster}, to produce text line proposals and character proposals.
Secondly, based on text line proposals, TDM predicts tight bounding boxes $B$~(\emph{i.e.}, rotated rectangles or polygons) and visual scores $S_{vis}$ for candidate text lines~(see Fig.~\ref{fig:arch}(d)).
Thirdly, using character proposals and bounding boxes $B$ as input, CRM recognizes the text contents $T$ of candidate text lines~(see Fig.~\ref{fig:arch}(e)).
After that, we feed the text contents $T$ to LM, to predict the linguistic scores $S_{lin}$ for candidate text lines.
Then we combine the visual scores $S_{vis}$ and linguistic scores $S_{lin}$ into the final scores $S$ as Eqn~\ref{eqn:s}.
\begin{equation}
	S = \lambda S_{vis} + (1 - \lambda)S_{lin},
	\label{eqn:s}
\end{equation}
where $\lambda$ is used to balance $S_{vis}$ and $S_{lin}$.
Finally, on the basis of the final scores $S$, we select the final bounding boxes and text contents of text lines~(see Fig.~\ref{fig:arch}(f)), by using NMS and removing low-score text lines.

The training process of AE TextSpotter is divided into two stages: 
1) training the backbone network, RPN, TDM, and CRM end-to-end; 
2) training LM with the predicted text lines and text contents.
In the first stage, we fix the weights of LM, and optimize the backbone network, RPN, TDM, and CRM by a joint loss of $\mathcal{L}_{rpn}$, $\mathcal{L}_{tdm}$, and $\mathcal{L}_{crm}$. 
In the second stage, we fix the weights of the backbone network, RPN, TDM, and CRM, and optimize LM by loss function $\mathcal{L}_{lm}$.

\subsection{Text Detection Module}
\begin{figure*}[t]
	\centering
	\setlength{\fboxrule}{0pt}
	\fbox{\includegraphics[width=0.85\textwidth]{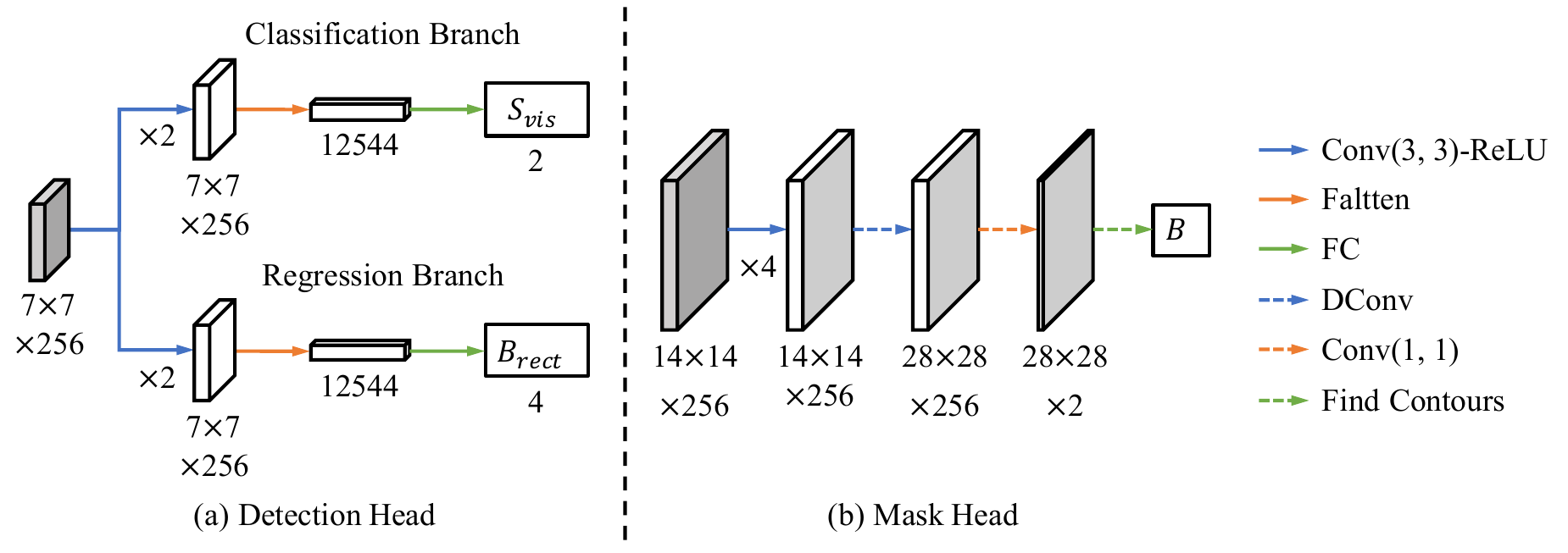}}
	\caption{The details of the detection head and the mask head in the text detection module~(TDM). ``Conv(3, 3)'', ``FC'' and ``DConv'' represent the convolution layer with the kernel size of 3$\times$3, fully connected layer, and deconvolution layer, respectively. ``$\times$2'' denotes a stack of two consecutive convolution layers.
	}
	\label{fig:tdm}
\end{figure*}
The text detection module~(TDM) aims to detect the candidate text lines with extremely high recall.
Due to the requirement of high recall, TDM needs to be implemented by a regression-based method, whose recall can be controlled by the classification score threshold and the NMS threshold.
In this work, following \cite{xie2019spcnet,liao2019maskpami,qin2019towards}, TDM is modified based on Mask R-CNN~\cite{he2017mask}. For each text line, it predicts an axis-aligned rectangular bounding box and the corresponding segmentation mask.

As shown in Fig.~\ref{fig:tdm}, there are a detection head and a mask head in TDM.
Firstly, we use RoIAlign~\cite{he2017mask} to extract feature patches~(7$\times$7$\times$256) for text line proposals predicted by RPN.
Secondly, we feed feature patches to the classification and regression branches in the detection head, to predict visual scores $S_{vis}$ and rectangular bounding boxes $B_{rect}$ of text lines.
Then, the new feature patches (14$\times$14$\times$256) are extracted for rectangular bounding boxes $B_{rect}$.
After that, using the new feature patches as input, the mask head predicts masks of text lines, and the mask contours are tight bounding boxes $B$.
Finally, we use a loose score threshold ${Thr}_{score}$ and a loose NMS threshold ${Thr}_{nms}$ to filter out redundant boxes in tight bounding boxes $B$, and the rest are candidate text lines.

\subsection{Character-based Recognition Module}
The character-based recognition module~(CRM) is applied to efficiently recognize a large number of candidate text lines produced by TDM.
CRM includes a detection head and a match-assemble~(MA) algorithm. The detection head runs only once for each input image to detect and classify characters;
and the proposed MA algorithm is used to group characters into text line transcriptions, which is rule-based and very fast. Therefore the CRM is very suitable for recognizing numerous candidate text lines predicted by TDM.

The detection head of CRM follows a pipeline similar to the detection head of TDM~(see Fig.~\ref{fig:tdm}~(a)). 
Due to the large number of character categories~(about 3,600 in IC19-ReCTS), the
detection head of CRM needs a better feature patch as input and a deeper classification
branch. Therefore, in the detection head of CRM, the size of feature patches extracted by RoIAlign is 14$\times$14$\times$256, and 
we change the number of convolutional layers in the classification branch to 4.
The output of the detection head is rectangular bounding boxes and contents of characters.

\begin{figure*}[t]
	\centering
	\setlength{\fboxrule}{0pt}
	\fbox{\includegraphics[width=0.9\textwidth]{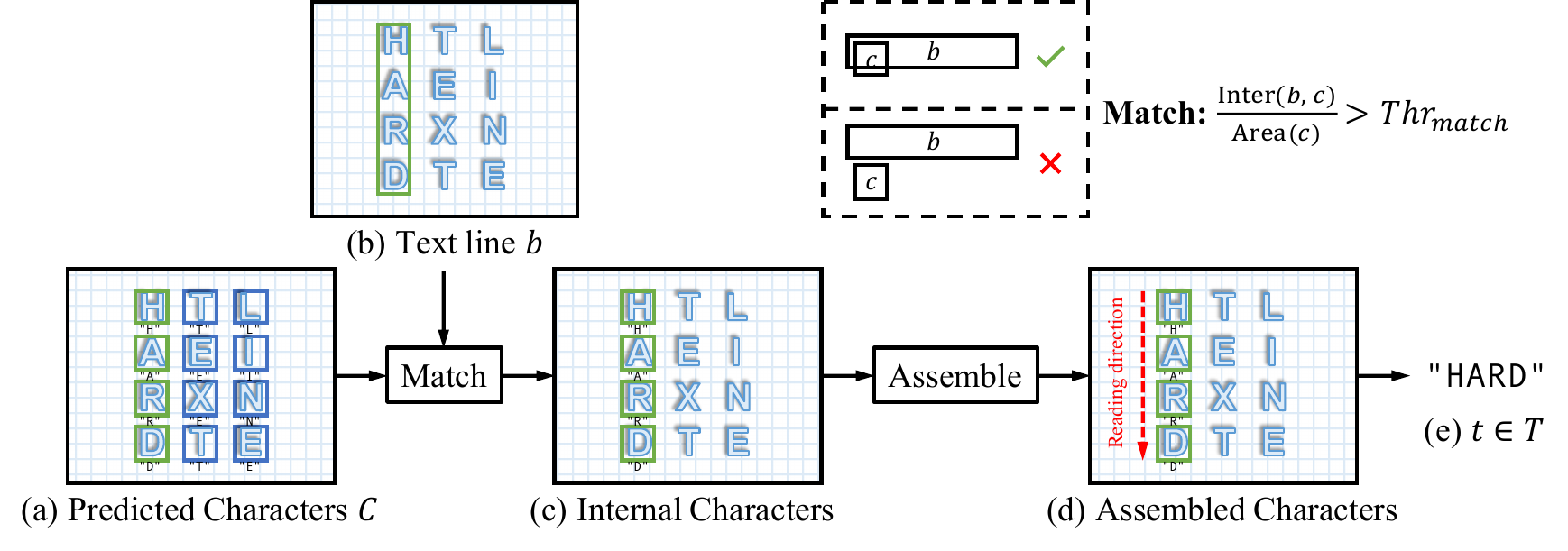}}
	\caption{The details of match-assemble~(MA) algorithm. ``Match'' is the operation of matching character boxes with a text line boxes. ``Assemble'' is the operation of assembling the internal characters to a text line transcription.
	}
	\label{fig:a2}
\end{figure*}

After character detection and classification, we propose a match-assemble~(MA) algorithm to obtain text contents.
The procedure of the MA algorithm is presented in Fig.~\ref{fig:a2}.
For each bounding box $b \in B$ of candidate text line,
we firstly \textbf{match} the predicted characters~(see Fig.~\ref{fig:a2}(a)) with the candidate text line $b$ by using Eqn.~\ref{eqn:c_in}, and get the internal characters $C_{in}$~(see Fig.~\ref{fig:a2}~(c)) for the candidate text line $b$.
\begin{equation}
	C_{in} = \{c \mid \frac{{\rm Inter}(c, b)}{{\rm Area}(c)} > Thr_{match} , c \in C\}
	\label{eqn:c_in}
\end{equation}
Here, ${\rm Inter}(\cdot)$, ${\rm Area}(\cdot)$ and $C$ are the intersection area of two boxes, the area of a box, and bounding boxes of detected characters, respectively.
Secondly, we \textbf{assemble} the internal characters $C_{in}$ according to their center positions.
Specifically, we arrange the characters from left to right, when the width of the external horizontal rectangle of the candidate text line is larger than its height, or else we arrange the characters from top to down.
After the arrangement, we concatenate character contents into the text content $T$ of the candidate text line.
\label{sec:CRM}

\subsection{Language Module}
The language module~(LM) is a language-based module that can re-score candidate text lines according to their contents.
Specifically, LM predicts a linguistic score to determine whether the text content follows the distribution of natural language.

\begin{figure*}[t]
	\centering
	\setlength{\fboxrule}{0pt}
	\fbox{\includegraphics[width=0.85\textwidth]{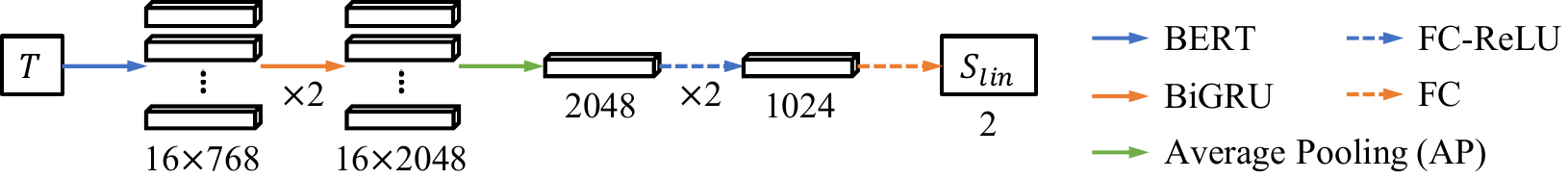}}
	\caption{The details of Language Module~(LM). ``BiGRU'', and ``FC'' represent the binary input gated recurrent unit~\cite{cho2014learning}, and fully connected layer, respectively. ``$\times$2'' denotes a stack of two consecutive layers.}
	\label{fig:lm}
\end{figure*}

The pipeline of LM is presented in Fig.~\ref{fig:lm}. At first, we use BERT~\cite{devlin2018bert}
, a widely-used pre-trained model in NLP, to extract sentence vectors ($16\times768$) for the input text content $T$. Here, 16 is the maximum length of text content, and we fill 0 for text lines whose length is less than 16.
Then we feed sentence vectors to a classification network composed of BiGRU, AP, and FC
, to predict the linguistic scores $S_{lin}$ of the input text contents. 
Here, BiGRU, AP, and FC represent binary input gated recurrent unit~\cite{cho2014learning}, average pooling layer, and fully connected layer, respectively.

The label generation for LM is based on Intersection over Union~(IoU) metric. Concretely, for every candidate text line, we consider the label of its text content to be positive if and only if there is a matched ground-truth whose IoU with the candidate text line is greater than a threshold ${IoU}_{pos}$. 

\subsection{Loss Function}
\label{sec:loss}
As mentioned in Sec.~\ref{sec:arch}, the training of AE TextSpotter has two stages.
Correspondingly, the loss function has two parts: 1) the visual loss function $\mathcal{L}_{vis}$; 2) the linguistic loss function $\mathcal{L}_{lm}$.

\textbf{The visual loss function} $\mathcal{L}_{vis}$ is used to optimize the backbone network, RPN, TDM, and CRM. It is a multi-task loss function that can be defined as Eqn.~\ref{eqn:l_vis}.
\begin{equation}
	\mathcal{L}_{vis} = \mathcal{L}_{rpn} + \mathcal{L}_{tdm} + \mathcal{L}_{crm}, 
	\label{eqn:l_vis}
\end{equation}
where $\mathcal{L}_{rpn}$ is loss function of RPN as defined in \cite{ren2015faster}. $\mathcal{L}_{tdm}$ and $\mathcal{L}_{crm}$ are loss functions of TDM and CRM respectively, which can be formulated as Eqn.~\ref{eqn:l_text} and  Eqn.~\ref{eqn:l_char}.
\begin{equation}
	\mathcal{L}_{tdm} = \mathcal{L}_{tdm}^{cls} + \mathcal{L}_{tdm}^{box} + \mathcal{L}_{tdm}^{mask}.
	\label{eqn:l_text}
\end{equation}
\begin{equation}
	\mathcal{L}_{crm} = \mathcal{L}_{crm}^{cls} + \mathcal{L}_{crm}^{box}.
	\label{eqn:l_char}
\end{equation}
Here, $\mathcal{L}_{tdm}^{cls}$ and $\mathcal{L}_{crm}^{cls}$ are loss functions of classification branches in TDM and CRM respectively, which are implemented by cross-entropy losses. $\mathcal{L}_{tdm}^{box}$ and $\mathcal{L}_{crm}^{box}$ are loss functions of regression branches in TDM and CRM respectively, which are implemented by smooth $L_1$~\cite{cho2014learning}. $\mathcal{L}_{tdm}^{mask}$ is the loss function of mask head in TDM, which is the same as the mask loss in \cite{he2017mask}.

\textbf{The linguistic loss function} $\mathcal{L}_{lm}$ is used to optimize LM. Actually, LM only involves a task of classifying the correct and incorrect text lines in candidates, so $\mathcal{L}_{lm}$ is implemented by a cross-entropy loss.

\section{Experiment}
\subsection{Datasets}
\label{sec:dataset}
AE TextSpotter needs character-level
annotations in the training phase. In addition text detection ambiguity is relatively rare in English due to its writing habit.
Therefore, we evaluate our method on a multi-language dataset with character-level annotations.

\textbf{ICDAR 2019 ReCTS}~(IC19-ReCTS)~\cite{zhang2019icdar} is a newly-released large-scale dataset that includes 20,000 training images and 5,000 testing images, covering multiple languages, such as Chinese, English and Arabic numerals. The images in this dataset are mainly collected in the scenes with signboards, 
All text lines and characters in this dataset are annotated with bounding boxes and transcripts.

\textbf{TDA-ReCTS} is a validation set for benchmarking text detection ambiguity, which contains 1,000 ambiguous images selected from the training set of IC19-ReCTS.
To minimize the impact of subjective factors, we designed rules to pick out images with the case of large character spacing or juxtaposed text lines, and then randomly sample 1,000 images among them as the validation set.

When selecting ambiguous images, an image is regarded as a sample with large character spacing, if at least one text line in the image has large character spacing. In addition, an image is treated as a sample with juxtaposed text lines, if there is a pair of text lines aligned with the top, bottom, left or right direction, and their characters have similar scales. More details about the both rules can be found in Appendix A.1.

\subsection{Implementation Details}
\label{sec:detail}
In the backbone network, the ResNet50~\cite{he2016deep} is initialized by the weights pre-trained on ImageNet~\cite{deng2009imagenet}.
In RPN, the aspect ratios of text line anchors are set to $\{$1/8, 1/4, 1/2, 1, 2, 4, 8$\}$ to match the extreme aspect ratios of text lines, and the aspect ratios of character anchors are set to $\{$1/2, 1, 2$\}$. All other settings in RPN are the same as in Mask R-CNN.
In TDM, the classification branch makes a binary classification of text and non-text, and the two post-processing thresholds~(\emph{i.e.}, ${Thr}_{score}$ and ${Thr}_{nms}$) in the regression branch are set to 0.01 and 0.9, respectively.
In CRM, the match threshold $Thr_{match}$ is set to 0.3.
In LM, candidate text lines are classified into correct and incorrect ones., and the training label threshold ${IoU}_{pos}$ is set to 0.8.
When combining the visual and linguistic scores, 
the balance weight $\lambda$ in Eqn.~\ref{eqn:s} is set to 0.7.
In the post-processing, the NMS threshold is set to 0.1 and the classification score threshold is set to 0.6.

As mentioned in Sec.~\ref{sec:arch}, the training of AE TextSpotter is divided into two phases. In the first phase, we train backbone network, RPN, TDM, and CRM with batch size 16 on 8 GPUs for 12 epochs, and the initial learning rate is set to 0.02 which is divided by 10 at 8 epoch and 11 epoch.
In the second phase, we fix the weights of the backbone network, RPN, TDM, and CRM, and train LM with batch size 16 on 8 GPUs for 12 epochs, and the initial learning rate is set to 0.2 which is divided by 10 at 8 epoch and 11 epoch.
We optimize all models using stochastic gradient descent (SGD) with a weight decay of $1 \times 10^{-4}$ and a momentum of 0.9. 

In the training phase, we ignore the blurred text regions labeled as ``DO NOT CARE'' in all datasets; and apply random scale, random horizontal flip, and random crop on training images. 

\subsection{Ablation Study}

\textbf{The proportion of ambiguous text lines.
}
Text detection ambiguity is caused by large character spacing and juxtaposed text lines. 
To count the proportion of the two problematic cases, we use the rules mentioned in Sec.~\ref{sec:dataset} to count the ambiguous text lines in the IC19-ReCTS training set and TDA-ReCTS.
As shown in Table~\ref{tab:p_tc}, ambiguous text lines occupy 5.97\% of total text lines in the IC19-ReCTS training set, which reveals that text detection ambiguity is not rare in the natural scene.
Moreover, the proportion of ambiguous text lines in TDA-ReCTS reaches 25.80\%.
Although the proportion in TDA-ReCTS is much larger than in the IC19-ReCTS training set, the normal text lines are still in the majority.


\begin{table*}[t]
	\centering
	\caption{The proportion of text lines with the problem of text detection ambiguity.}
	\setlength{\tabcolsep}{1.2mm}
	\begin{tabular}{l|c|c|c|c}
	\hline
	\multirow{2}{*}{Type} & \multicolumn{2}{c|}{IC19-ReCTS Training Set} & \multicolumn{2}{c}{TDA-ReCTS} \\
	\cline{2-5}
	& Number & Proportion & Number & Proportion \\
	\hline
	Large Character Spacing & 1,241 & 1.14\% & 589 & 7.07\% \\
	Juxtaposed Text Lines & 5,379 & 4.94\% & 1,615 & 19.39\% \\
	Union of two Categories & 6,509 & 5.97\% & 2,148 & 25.80\% \\
	\hline
	All & 108,963 & - & 8,327 & - \\
	\hline
\end{tabular}
	\label{tab:p_tc}
\end{table*}

\textbf{The high recall of the text detection module.} 
To verify the high recall of the text detection module (TDM), we compare the recall and the number of candidate text lines per image on TDA-ReCTS, under different visual score thresholds ${Thr}_{score}$ and the NMS thresholds ${Thr}_{nms}$.
As shown in Table.~\ref{tab:recall}, when ${Thr}_{score}=0.0$ and ${Thr}_{nms}=1.0$, the recall of TDM reaches the upper bound of 0.971.
However, in this case, the number of candidate text lines per image is 971.5, which is too large to be recognized.
When ${Thr}_{score}=0.01$ and ${Thr}_{nms}=0.9$, the recall of TDM is 0.969, slightly lower than the upper bound, but the number of candidate text lines per image is reduced to 171.7.
If we use common post-processing thresholds (\emph{i.e.}, ${Thr}_{score}=0.5$ and ${Thr}_{nms}=0.5$), the recall of TDM is only 0.852, which will result in many missed detection of text lines.
Therefore, in experiments, the values of ${Thr}_{score}$ and ${Thr}_{nms}$ are set to 0.01 and 0.9 by default, which can achieve a high recall without too many candidate text lines.

\textbf{The effectiveness of the character-based recognition module.}
The character-based recognition module (CRM) aims to fast recognize numerous candidate text lines predicted by TDM. To test the speed of CRM, we compare the time cost per text line of CRM and mainstream recognizers~(\emph{e.g.} CRNN~\cite{shi2016crnn} and ASTER~\cite{shi2018aster}). For a fair comparison, we run these methods on 1080Ti GPU and report their time cost per image on the IC19-ReCTS test set. Note that, the input text images of CRNN and ASTER are scaled to a height of 32 when testing.
In AE TextSpotter, the recognition module is used to recognize the candidate text lines, whose number reaches \textbf{171.7 per image}.
Table.~\ref{tab:ctrm_speed} shows the speed of different methods. CRNN and ASTER cost 966.7 ms and 4828.2 ms respectively to recognize \textbf{all candidate text lines} in each image.
The time cost of CRM on each image is only 122.3 ms, so CRM runs much faster than CRNN and ASTER.
In addition, using the same detection result predicted by AE TextSpotter, our CRM can achieve the 1-NED of 71.81 on the test set of IC19-ReCTS, which is better than CRNN and ASTER.

\makeatletter
\newcommand\figcaption{\def\@captype{figure}\caption}
\newcommand\tabcaption{\def\@captype{table}\caption}
\makeatother
\begin{figure}[t]
	\begin{minipage}[t]{.58\linewidth}
		\tabcaption{The recall of TDM and the number of candidate text lines per image under different post-processing thresholds.}
		\centering
		\setlength{\tabcolsep}{0.8mm}
		\begin{tabular}{c|c|c|c}
	\hline
	$Thr_{score}$ & $Thr_{nms}$ & Recall & Number of Candidates \\
	\hline
	0 & 1 & 0.971 & 971.5 \\
	0.01 & 0.9 & 0.969 & 171.7 \\
	0.5 & 0.5 & 0.852 & 11.1 \\
	\hline
\end{tabular}

		\label{tab:recall}
	\end{minipage}
	\begin{minipage}[t]{.4\linewidth}
		\tabcaption{The time cost per image and 1-NED of different recognizers on the IC19-ReCTS test set. 
		}
		\centering
		\setlength{\tabcolsep}{0.8mm}
		\begin{tabular}{l|c|c}
	\hline
	Method & Time Cost & 1-NED \\
	\hline
	CRNN~\cite{shi2016crnn} & 966.7~ms & 68.32 \\
	ASTER~\cite{shi2018aster} & 4828.2~ms & 58.49 \\
	\hline
	CRM~(ours) & 122.3~ms & 71.81 \\
	\hline
\end{tabular}

		\label{tab:ctrm_speed}
	\end{minipage}
\end{figure}

\begin{figure*}[t]
	\centering
	\setlength{\fboxrule}{0pt}
	\fbox{\includegraphics[width=0.9\textwidth]{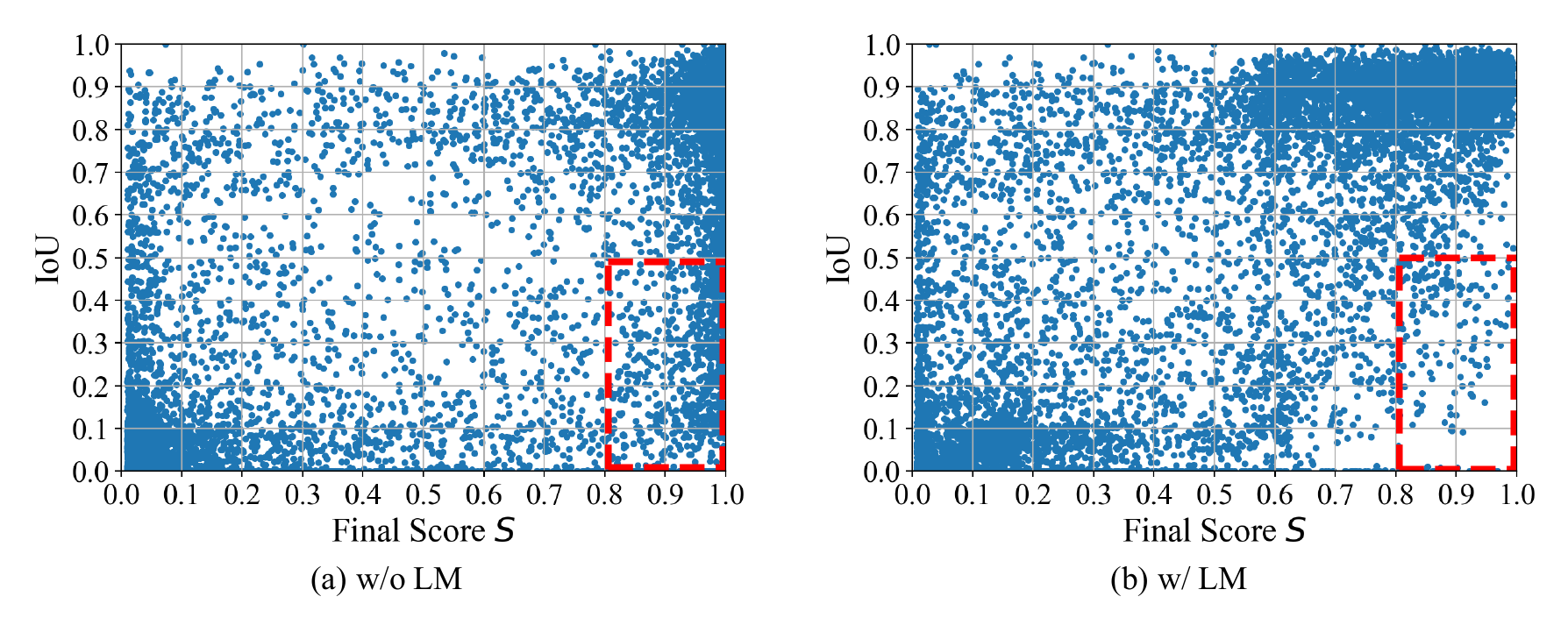}}
	\caption{The distributions of detection results predicted by models with and without LM. The points in \textbf{red dotted boxes} are false positives with high scores.
	}
	\label{fig:si}
\end{figure*}

\textbf{The effectiveness of the language module.} To investigate the effectiveness of the language module (LM), we compare the performance of the models with and without LM on TDA-ReCTS.
For a fair comparison, we build a model without LM by removing LM from AE TextSpotter in the training phase and setting $\lambda$ in Eqn.~\ref{eqn:s} to 1.0 in the testing phase.
As shown in Table \ref{tab:tc_rects}, on TDA-ReCTS, 
the model with LM obtains the F-measure of 81.39\% and the 1-NED of 51.32\%, significantly surpassing the model without LM by 3.46\% and 3.57\%.

In addition, we analyze the quality of the detection results predicted by the models with and without LM on TDA-ReCTS.
Generally, the quality of a bounding box can be measured by its IoU with the matching ground truth, and a bounding box is considered to be correct if its IoU exceeds 0.5.
As shown in Fig.~\ref{fig:si}~(a), the detection results of the model without LM have many incorrect results (IoU $<$ 0.5), even though their final scores are greater than 0.8.
However, this problem is alleviated in the model with LM~(see Fig.~\ref{fig:si}~(b)), where there are few detection results with high scores but low IoU.
These experiment results demonstrate the power of LM.

\subsection{Comparisons with State-of-the-Art Methods}
To show the effectiveness of the proposed AE TextSpotter, we compare it with state-of-the-art methods on TDA-ReCTS and IC19-ReCTS test set.

On TDA-ReCTS, we compare the proposed AE TextSpotter with some representative methods.
All results are obtained by using authors provided or officially released codebases,
and these methods are trained on the 19,000 training images in IC19-ReCTS~(except 1,000 images in TDA-ReCTS).
In the testing phase, we scale the short side of images to 800, and evaluate the performance of these methods on TDA-ReCTS by the evaluation metric defined in \cite{zhang2019icdar}.
Table \ref{tab:tc_rects} shows the performance of these methods. The proposed AE TextSpotter achieves the F-measure of 81.39\% and the 1-NED of 51.32\%, outperforming the counterparts by at least 4.01\% and 4.65\% respectively.
Here, 1-NED denotes normalized edit distance, a recognition metric defined in \cite{zhang2019icdar}.
Note that, the proportion of ambiguous text lines in TDA-ReCTS is only 25.80\%, and we believe the improvement will be more significant if the proportion is larger.

On the IC19-ReCTS test set, we train models on the complete training set of IC19-ReCTS.
As shown in Table~\ref{tab:rects}, our AE TextSpotter achieves the F-measure of 91.80\% and the 1-NED of 71.81\%, surpassing other methods.
Moreover, when we enlarge the scale of input images (short size 1000), the F-measure and 1-NED of our method can be boosted to 91.86\% and 73.12\%.
Without extra datasets, the single-scale detection performance of our method is comparable to the methods that use extra datasets, multi-scale testing, and model ensemble (91.86\% \emph{v.s.} 92.99\%).
Unlike the detection task, the multi-language recognition task relies more heavily on extra datasets.
Therefore, our recognition performance is lower than the methods using extra datasets, multi-scale testing, and ensemble.

\begin{figure*}[t]
	\centering
	\setlength{\fboxrule}{0pt}
	\fbox{\includegraphics[width=0.98\textwidth]{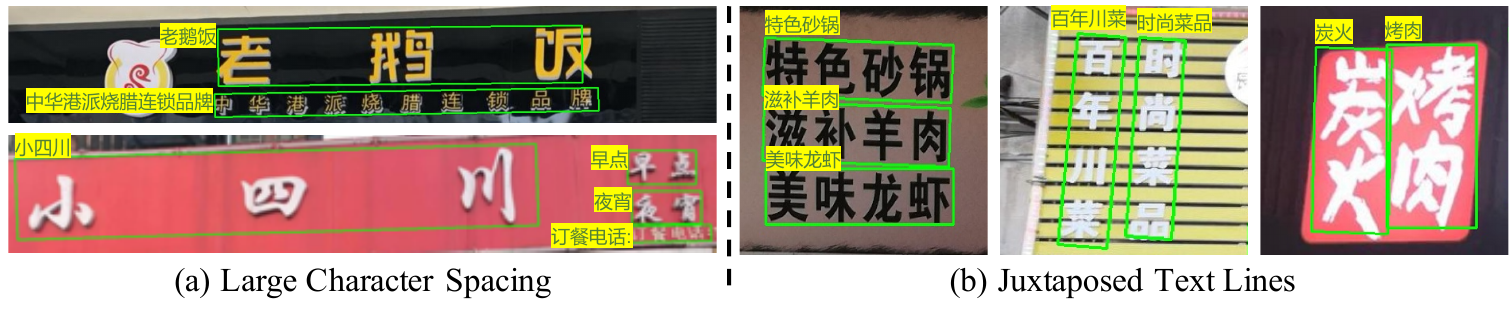}}
	\caption{Visualization results of AE TextSpotter on TDA-ReCTS.
	}
	\label{fig:res}
\end{figure*}

These results demonstrate that the proposed AE TextSpotter can correctly detect and recognize multi-language text lines, even in ambiguous scenarios. 
Some qualitative results on TDA-ReCTS are shown in Fig. \ref{fig:res}.

\begin{table*}[t]
	\centering
	\caption{The single-scale results on TDA-ReCTS. ``P'', ``R'', ``F'' and ``1-NED''  mean the precision, recall, F-measure, and normalized edit distance~\cite{zhang2019icdar}, respectively. 
	}
	\setlength{\tabcolsep}{1.2mm}
	\begin{tabular}{l|c|c|c|c|c|c}
	\hline
	\multirow{2}{*}{Method} & \multirow{2}{*}{Venue} & \multirow{2}{*}{Backbone} &  \multicolumn{4}{c}{TDA-ReCTS} \\
	\cline{4-7}
	& & & P & R & F & 1-NED  \\
	\hline
	EAST~\cite{zhou2017east} & CVPR'17 & VGG16 & 70.58 & 61.59 & 65.78 & - \\
	PSENet~\cite{wang2019shapecvpr} & CVPR'19 & ResNet50 & 72.14 & 65.53 & 68.68 & - \\
	FOTS~\cite{liu2018fots} & CVPR'18 & ResNet50 & 68.20 & 70.00 & 69.08 & 34.87 \\
	Mask TextSpotter~\cite{liao2019maskpami} & TPAMI & ResNet50 & 80.07 & 74.86 & 77.38 & 46.67 \\
	\hline
	Detection Only & - & ResNet50 & 79.58 & 74.93 & 77.19 & - \\
	AE TextSpotter w/o LM & - & ResNet50 & 79.18 & 76.72 & 77.93 & 47.74 \\
	AE TextSpotter & - & ResNet50 & \textbf{84.78} & \textbf{78.28} & \textbf{81.39} &  \textbf{51.32} \\
	\hline
\end{tabular}

	\label{tab:tc_rects}
\end{table*}

\begin{table*}[t]
	\centering
	\caption{The single-scale results on the IC19-ReCTS test set. ``P'', ``R'', ``F'' and ``1-NED'' represent the precision, recall, F-measure, and normalized edit distance, respectively. ``*'' denotes the methods in competition~\cite{zhang2019icdar}, which use extra datasets, multi-scale testing, and model ensemble. ``800$\times$'' means that the short side of input images is scaled to 800.
	}
	\setlength{\tabcolsep}{1.2mm}
	\begin{tabular}{l|c|c|c|c|c|c}
	\hline
	\multirow{2}{*}{Method} & \multirow{2}{*}{Venue} & \multirow{2}{*}{Backbone} &  \multicolumn{4}{c}{IC19-ReCTS Test Set} \\
	\cline{4-7}
	& & & P & R & F & 1-NED  \\
	\hline
	Tencent-DPPR Team*~\cite{zhang2019icdar} & Competition & - & 93.49 & 92.49 & 92.99 & 81.45 \\
	SANHL\_v1*~\cite{zhang2019icdar} & Competition & - & 91.98 & 93.86 & 92.91 & 81.43 \\
	HUST\_VLRGROUP*~\cite{zhang2019icdar} & Competition & - & 91.87 & 92.36 & 92.12 & 79.38 \\
	\hline
	\hline
	EAST~\cite{zhou2017east} & CVPR'17 & VGG16 & 74.25 & 73.65 & 73.95 & - \\
	PSENet~\cite{wang2019shapecvpr} & CVPR'19 & ResNet50 & 87.31 & 83.86 & 85.55 & - \\
	FOTS~\cite{liu2018fots} & CVPR'18 & ResNet50 & 78.25 & 82.49 & 80.31 & 50.83 \\
	Mask TextSpotter~\cite{liao2019maskpami} & TPAMI & ResNet50 & 89.30 & 88.77 & 89.04 & 67.79 \\
	\hline
	Detection Only & - & ResNet50 & 90.72 & 90.73 & 90.72 & - \\
	AE TextSpotter w/o LM & - & ResNet50 & 91.54 & 90.78 & 91.16  & 70.95 \\
	AE TextSpotter (800$\times$) & - & ResNet50 & 92.60 & 91.01 & 91.80 & 71.81 \\
	AE TextSpotter (1000$\times$) & - & ResNet50 & \textbf{92.28} & \textbf{91.45} & \textbf{91.86} & \textbf{73.12}\\
	\hline
\end{tabular}

	\label{tab:rects}
\end{table*}


\subsection{Time Cost Analysis of AE TextSpotter}
We analyze the time cost of all modules in AE TextSpotter. Specifically, we evaluate our methods on images in TDA-ReCTS and
calculate the average time cost per image. All results are tested by PyTorch~\cite{paszke2019pytorch} with the batch size of 1 on one 1080Ti GPU in a single thread.
Table~\ref{tab:speed} shows the time cost of backbone, TDM, CRM, LM, and post-processing respectively. Among them, the detection module takes up most of the execution time, because it needs to predict the mask for every candidate text line. The entire framework can run 1.3 FPS when the short size of input images are 800.

\begin{table*}[t]
	\centering
	\caption{The time cost of all modules in AE TextSpotter.
	}
	\setlength{\tabcolsep}{1.2mm}
	\begin{tabular}{c|c|c|c|c|c|c|c}
	\hline
	\multirow{2}{*}{Scale} & \multirow{2}{*}{FPS} & \multicolumn{6}{c}{Time Cost per Image (ms)} \\
	\cline{3-8}
	 & & Backbone & RPN & TDM & CRM & LM & Post-proc\\
	\hline
	800$\times$ & 1.3 & 31.3 & 35.1 & 440.0 & 122.3  & 118.6 & 3.8 \\
	1000$\times$ & 1.2 & 63.1 & 42.6 & 487.2 & 127.8 & 121.1 & 3.9  \\
	\hline
\end{tabular}

	\label{tab:speed}
\end{table*}

\subsection{Discussion}
As demonstrated in previous experiments, the proposed
AE TextSpotter works well in most cases, including scenarios of text detection ambiguity. Nevertheless, there is still room for improvement in some parts as follows.
\textbf{1)} The recognition performance of our method is not high (only 51.32\% in Table~\ref{tab:tc_rects}), supressing the effectiveness of LM. 
\textbf{2)} TDA-ReCTS is selected by rules that only give a vague description of ambiguity in text detection. Therefore, a manually selected dataset is important for further exploring text detection ambiguity. 
\textbf{3)} The proposed AE TextSpotter is a baseline for removing text detection ambiguity, and it is still worthwhile to further explore a better way to combine visual and linguistic representation.

Although our AE TextSpotter works well in most cases. It still fails in some intricate images, such as text-like regions, strange character arrangement, and texts in fanc styles. More details about these failed cases can be found in Appendix A.2.


\section{Conclusion \& Future Work}
In this paper, we first conducted a detailed analysis of the ambiguity problem in text detection, and revealed the importance of linguistic representation in solving this problem.
Then we proposed a novel text spotter, termed AE TextSpotter, which introduces linguistic representation to eliminate ambiguity in text detection. 
For the first time, linguistic representation is utilized in scene text detection to deal with the problem of text detection ambiguity. 
Concretely, we propose the Language Module~(LM), which can learn to re-score the text proposals by linguistic representation. LM can effectively lower the scores of incorrect text lines while improve the scores of correct proposals.
Furthermore, a new validation set is proposed for benchmarking the ambiguity problem.
Extensive experiments demonstrate the advantages of our method, especially in scenarios of text detection ambiguity.

In the future, we will strive to build a more efficient and fully end-to-end network for ambiguous text spotting, and plan to build a large-scale and more challenging dataset which is full of ambiguous text samples.

\section*{Acknowledgments}
This work is supported by the Natural Science Foundation of China under Grant 61672273 and Grant 61832008, the Science Foundation for Distinguished Young Scholars of Jiangsu under Grant BK20160021, and Scientific Foundation of State Grid Corporation of China (Research on Ice-wind Disaster Feature Recognition and Prediction by Few-shot Machine Learning in Transmission Lines).

	\clearpage
	%
	%
	\bibliographystyle{splncs04}
	\bibliography{egbib}

	\section{Appendix}
	\subsection{Ambiguous Images Selection}
	To keep the objectivity of TDA-ReCTS validation set, we designed two rules to select ambiguous images from the training set of IC19-ReCTS~\cite{zhang2019icdar}, and then randomly sample 1,000 images among them as the validation set.
	
	For an image, we consider its text lines as $L$, and the internal characters of the $i$th text line as $C_{in}^i$.
	
	An image is regarded as a sample with large character spacing, if at least one text line in the image has large character spacing. The character spacing of a text line $\ell_i \in L$ is large, if it satisfies Eqn.~\ref{eqn:r_large}.
	\begin{equation}
	\frac{\sum_{c_j^i \in C_{in}^i} {\rm min}_{k \ne j, c_k^i \in C_{in}^i}\mathcal{D}_{c}(c_j^i, c_k^i)}{\sum_{c_j^i \in C_{in}^i} {\rm Scale}(c_j^i)} > 2,
	\label{eqn:r_large}
	\end{equation}
	where $c_j^i$ means the $j$th character in the internal characters $C_{in}^i$. $\mathcal{D}_{c}(\cdot)$ denotes the Euclidean distance between the center points of two character boxes. ${\rm Scale}(\cdot)$ signifies the scale of a character box, which is calculated by the square root of the box area.
	
	Moreover, an image is considered as a sample with juxtaposed text lines, if the image has a pair of text lines aligned to the top, bottom, left or right direction, and characters in them have similar scales. Two text lines~(\emph{i.e.}, $\ell_i$ and $\ell_j$) are aligned, if they satisfy Eqn.~\ref{eqn:r_multi1}. The characters in two text lines~(\emph{i.e.}, $\ell_i$ and $\ell_j$) have similar scales, if they satisfy Eqn.~\ref{eqn:r_multi2}.
	\begin{equation}
	\frac{| C_{in}^i | {\rm min}_{d\in\{t, b, l, r\}}\mathcal{D}_d(\ell_i, \ell_j)}{\sum_{c_k^i \in C_{in}^i} {\rm Scale}(c_k^i)} < \frac{1}{10}.
	\label{eqn:r_multi1}
	\end{equation}
	\begin{equation}
	\frac{9}{10} \le \frac{| C_{in}^j | \sum_{c_k^i \in C_{in}^i} {\rm Scale}(c_k^i)}{| C_{in}^i | \sum_{c_k^j \in C_{in}^j} {\rm Scale}(c_k^j)} \le \frac{10}{9}.
	\label{eqn:r_multi2}
	\end{equation}
	Here, $| C_{in}^i |$ is the number of internal characters of $i$th text line. $\mathcal{D}_{d=\{t, b, l, r\}}(\cdot)$ represents the absolute difference of the \textbf{t}op/\textbf{b}ottom/\textbf{l}eft/\textbf{r}ight of two text line boxes.
	\subsection{Failure Analysis}
\begin{figure*}[h]
		\centering
		\setlength{\fboxrule}{0pt}
		\fbox{\includegraphics[width=0.98\textwidth]{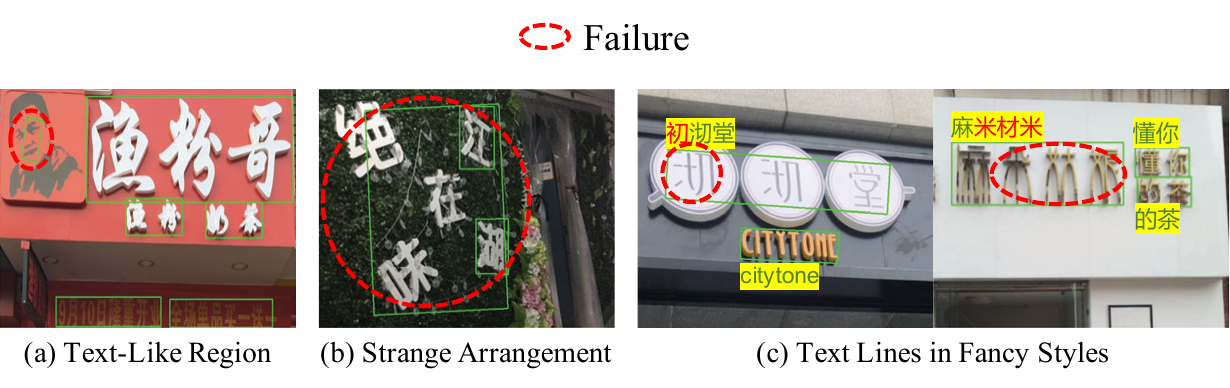}}
		\caption{Failed examples.}
		\label{fig:fail}
	\end{figure*}

	As demonstrated in previous experiments, the proposed
AE TextSpotter works well in most cases. It still fails in some intricate images, such as text-like regions (see Fig.~\ref{fig:fail}(a)), strange character arrangement~(see Fig.~\ref{fig:fail}(b)), and texts in fancy styles~(see Fig.~\ref{fig:fail}(c)). 
Most text-like regions can be removed with a high classification score threshold, but there are still some inevitable error detections.
In Fig.~\ref{fig:fail}(b), characters are arranged in ``x'' shape, which does not follow common writing habit. Therefore, it is difficult to detect text lines with extremely strange character arrangements.
The detection of text lines in fancy styles is feasible.
However, due to fancy styles, it is hard to correctly recognize text contents in these text lines.

\subsection{Visual Comparisons between AE TextSpotter and State-of-The-Art Methods.}
\begin{figure*}[h]
	\centering
	\setlength{\fboxrule}{0pt}
	\fbox{\includegraphics[width=0.98\textwidth]{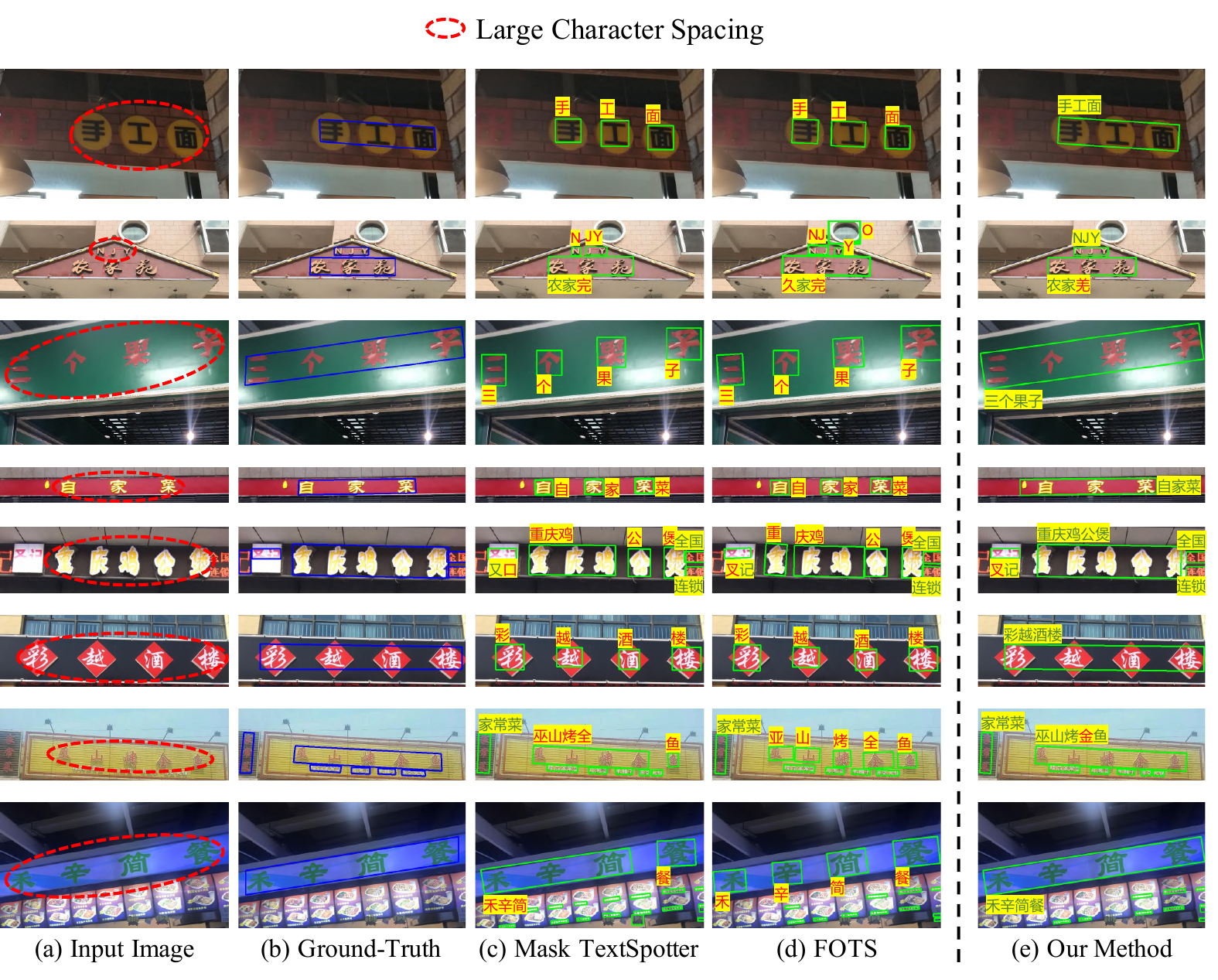}}
	\caption{Examples of text lines with large character spacing. (a) are original images. (b) are ground-truths. (c) are results of Mask TextSpotter~\cite{liao2019maskpami}. (d) are results of FOTS~\cite{liu2018fots}. (e) are results of our AE TextSpotter.}
	\label{fig:res_lcs}
\end{figure*}

\begin{figure*}[h]
	\centering
	\setlength{\fboxrule}{0pt}
	\fbox{\includegraphics[width=0.98\textwidth]{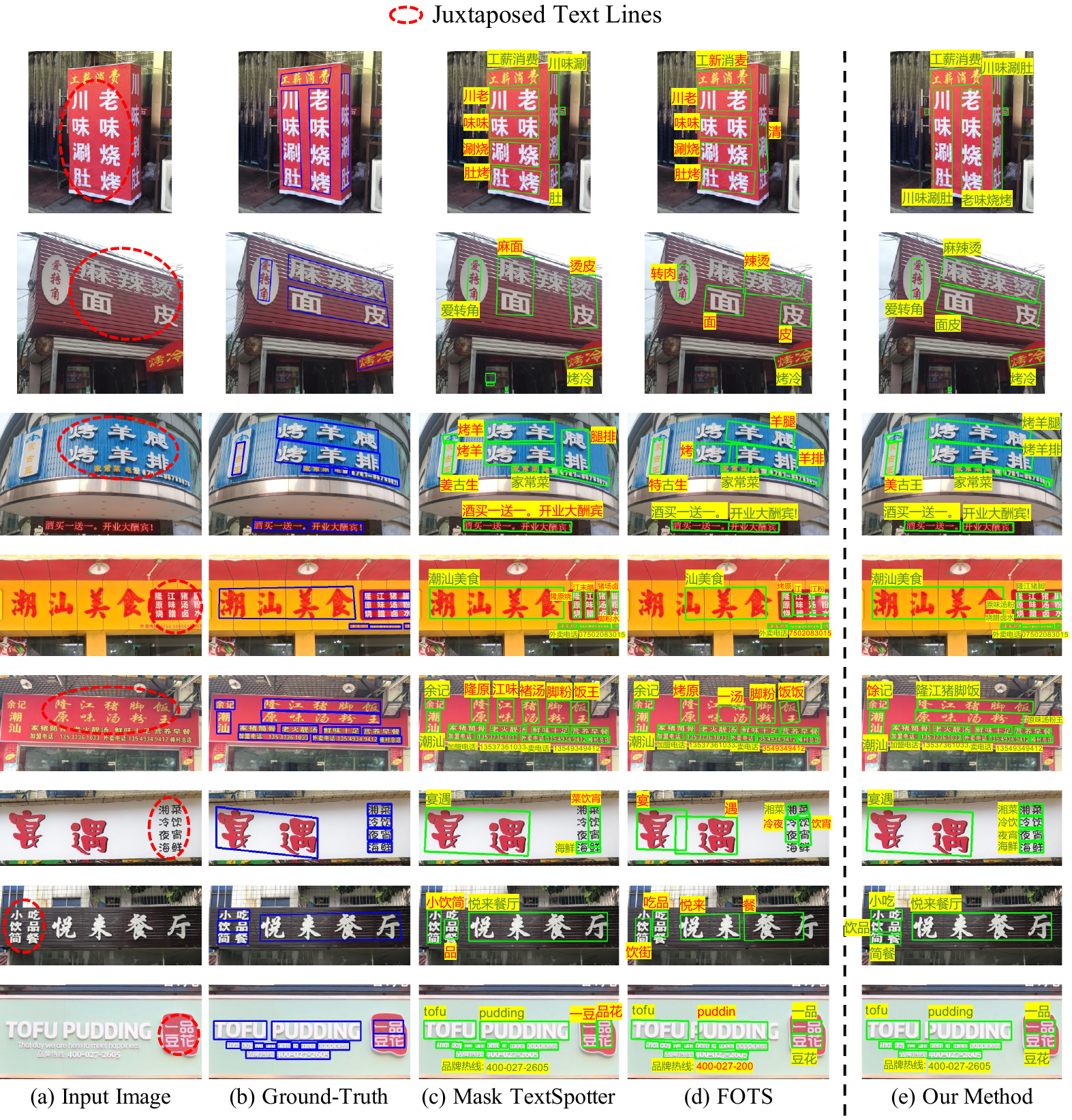}}
	\caption{Examples of juxtaposed text lines. (a) are original images. (b) are ground-truths. (c) are results of Mask TextSpotter~\cite{liao2019maskpami}. (d) are results of FOTS~\cite{liu2018fots}. (e) are results of our AE TextSpotter.}
	\label{fig:res_jtl}
\end{figure*}

In this section, we present the effectiveness of our AE TextSpotter by comparing visual results of different methods.
Specifically, we visualize and analyse the results predicted by Mask TextSpotter~\cite{liao2019maskpami}, FOTS~\cite{liu2018fots}, and  our AE TextSpotter on TDA-ReCTS.
For a fair comparison, both methods are trained on the 19,000 training
images in IC19-ReCTS~\cite{zhang2019icdar} (except 1,000 images in TDA-ReCTS). In the testing
phase, we scale the short side of test images to 800.
Fig.~\ref{fig:res_lcs} and Fig.~\ref{fig:res_jtl} show the examples of large character spacing and juxtaposed text lines, respectively.
In these examples, our method is clearly better than Mask TextSpotter and FOTS.
From these results, we can find that the proposed AE TextSpotter has the following abilities.
\begin{itemize}
\item Detecting and recognizing text lines with large character spacing;
\item Detecting and recognizing juxtaposed text lines;
\item Detecting and recognizing the text lines with various orientations;
\item Detecting and recognizing the multi-language~(\emph{e.g.} English, Chinese and Arabic numerals) text lines;
\item Thanks to the strong feature representation, our method is also robust to complex and unstable illumination, different colors and variable scales.
\end{itemize}

\end{document}